\documentclass{ifacconf}

\usepackage{graphicx}      
\usepackage{natbib}        
\usepackage{amsmath}
\usepackage{multirow}
\usepackage{booktabs}
\usepackage{caption}
\usepackage{subcaption}
\begin{document}
\begin{frontmatter}

\title{Computer Vision Approaches for Automated Bee Counting Application} 

\thanks[footnoteinfo]{The completion of this paper was made possible by grant No. FEKT-S-23-8451 - "Research on advanced methods and technologies in cybernetics, robotics, artificial intelligence, automation and measurement" financially supported by the Internal science fund of Brno University of Technology.}

\author[First]{Simon Bilik} 
\author[Second]{Ilona Janakova} 
\author[Second]{Adam Ligocki}
\author{Dominik Ficek}
\author[Second]{Karel Horak}

\address[First]{Department of Control and Instrumentation, Faculty of Electrical Engineering and Communication, Brno University of Technology, Brno, Czech Republic \textbf{and} Computer Vision and Pattern Recognition Laboratory, Department of Computational Engineering, Lappeenranta-Lahti University of Technology LUT, Lappeenranta, Finland (e-mail: bilik@vut.cz).}
\address[Second]{Department of Control and Instrumentation, Faculty of Electrical Engineering and Communication, Brno University of Technology, Brno, Czech Republic (e-mail: janakova@vut.cz, ligocki@vut.cz, horak@vut.cz).}

\begin{abstract}                
Many application from the bee colony health state monitoring could be efficiently solved using a computer vision techniques. One of such challenges is an efficient way for counting the number of incoming and outcoming bees, which could be used to further analyse many trends, such as the bee colony health state, blooming periods, or for investigating the effects of agricultural spraying. In this paper, we compare three methods for the automated bee counting over two own datasets. The best performing method is based on the ResNet-50 convolutional neural network classifier, which achieved accuracy of 87\% over the BUT1 dataset and the accuracy of 93\% over the BUT2 dataset.
\end{abstract}


\end{frontmatter}

\section{Introduction}

The automated bee counting based on computer vision techniques was solved with numerous methods involving the conventional computer vision approach, deep learning classification and object detection as extensively described in~\cite{BILIK2024108560}. The current trends in this field show a decline of the conventional computer vision techniques in favor of the object detectors and CNN-based classifiers. Nevertheless, the conventional computer vision solutions are still more explainable and might be more computationally efficient, which is important especially on the embedded devices and testers.

In this work, we compare three approaches for the automated bee counting based on the computer vision techniques. The first solution is our own, based on conventional computer vision-based motion signal analysis. It is applicable on an embedded device, allowing real-time processing at a 5 Hz sampling frequency. The second proposed approach is based on deep learning classifiers, and finally, the third proposed approach utilizes an object detector technique. In the conclusions, we compare the achieved results and possible improvements to the proposed methods.

\section{Related work}

The research task is analyzing the movement of individual objects, such as bees traversing through a predefined environment. Numerous existing techniques tackle this problem, each with its own pros and cons. In this chapter, we briefly discuss the practical applicability of several techniques to our task.

With the uprising of convolutional neural networks and specifically object detectors many works focus on tracking objects with direct access to the detector's predictions on each frame, for example, the Deep SORT \cite{8296962} algorithm. Another motion analysis technique is to use the initial state of objects in their first appearance in an analyzed sequence to track them through the sequence using just visual information. A common approach for this problem is to use Siamese neural networks \cite{Li_2019_CVPR} that formulate the problem as convolutional feature cross-correlation between objects and regions of interest. Again this approach requires the use of computationally expensive neural networks. Furthermore, due to the large bee traffic through the observed alley, the initialization step would've had to be run on every time step, increasing the computational cost even more.

Some works focus on analyzing temporal information through a low-dimensional subspace of ambient space with clustering techniques, known as subspace clustering techniques \cite{6482137} and are used to separate low-dimensional data to their parent subspaces. This approach seems feasible for our needs at first glance but is not applicable for numerous reasons. The most common problem with subspace clustering variants is that they set requirements on the low-dimensional data that are not realistic for us to meet; for example, they require prior knowledge of several subspaces \cite{1211332,5457695} or prior knowledge of some data points each of the subspaces contains \cite{10.1007/978-3-540-30212-4_2}. Other approaches generally rely on building a similarity matrix from input data matrix \cite{10.1007/11744085_8, 4270260, NIPS2001_801272ee}.

The computer vision based bee counting have been described in numerous papers. The conventional approach could be presented by numerous papers including for example the 3D tracking techniques \cite{Chiron2013}, image velocimetry algorithm integrated with a RaspberryPi 3 based system \cite{app10062042} supplemented also with the sensor data in \cite{kulyukin2022integration}. The CNN based solutions are represented in a smaller count, for example with the paper \cite{app9183743} describing a similar system as in \cite{app10062042}, but using the CNN classifier to detect and count the bees. An alternative to this approach is described in \cite{10.3389/fcomp.2021.769338}, where the authors detect parts of the bee's body in order to perform its tracking. The most popular approach seems to be the object detection represented by papers \cite{ryu2021honeybee} utilizing the YoloV4 and device with tunnels allowing the bees to pass through only in one direction, followed by the YoloV5 video analysis presented in \cite{YOLO_Vcely}, where the authors investigate bee traffic in front of the hive. A similar approach using the SSD is presented in \cite{9788643}.

\section{Experimental description}

\subsection{Dataset description}

For our experiments, we used two datasets containing bees incoming and leaving the beehive through narrow passages inspired by~\cite{CHEN2012100} and~\cite{BJERGE2019104898}, which should mechanically separate the bees for their easier image processing. The first dataset BUT1 contains 1153 images with 2737 bee annotations in total and it is available online in~\cite{BeeDS1}. Because of the unsuitable illumination and background of the BUT1 dataset, the second dataset BUT2 was created as an annotated subset from three captured days from the improved dataset~\cite{BeeDS2} and it contains 2154 images with 8185 annotations. The annotated variant is available upon request and the whole dataset is shown in Table~\ref{tb:Datasets}.

\begin{table*}[ht!]
    \begin{center}
    \caption{Overview of the used datasets}\label{tb:Datasets}
    \begin{tabular}{ccrrrrrrr}
    \multicolumn{3}{c}{} & \multicolumn{6}{c}{Annotation class} \\
    \multicolumn{2}{c}{Dataset}   & \multicolumn{1}{l}{Images} & \multicolumn{1}{c}{bee\_abdomen} & \multicolumn{1}{c}{bee\_cluster} & \multicolumn{1}{c}{bee\_complete\_in} & \multicolumn{1}{c}{bee\_complete\_out} & \multicolumn{1}{c}{bee\_head} & \multicolumn{1}{c}{Total} \\
    \toprule
    \multirow{4}{*}{BUT1} & Train & 753   & 534  & 146  & 658  & 27   & 338  & 1703 \\
                          & Valid & 200   & 161  & 51   & 127  & 45   & 83   & 467  \\
                          & Test  & 200   & 142  & 113  & 182  & 24   & 106  & 567  \\
                          & Total & 1153  & 837  & 310  & 967  & 96   & 527  & 2737 \\
    \midrule
    \multirow{4}{*}{BUT2} & Train & 1754  & 586  & 525  & 855  & 701  & 929  & 3596 \\
                          & Valid & 200   & 124  & 100  & 107  & 218  & 169  & 718  \\
                          & Test  & 200   & 578  & 617  & 590  & 1311 & 775  & 3871 \\
                          & Total & 2154  & 1288 & 1242 & 1552 & 2230 & 1873 & 8185 \\             
    \hline
    
    \end{tabular}
    \end{center}
\end{table*}

\begin{figure}[ht!]
    \begin{center}
    \includegraphics[width=8.4cm]{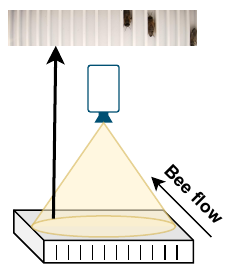}    
    \caption{Illustration of the measurement setup.} 
    \label{fig:MSetup}
    \end{center}
\end{figure}

\begin{figure*}[htp]
	\centering
    \begin{subfigure}[ht!]{0.9\textwidth}
        \centering
        \includegraphics[width=\textwidth]{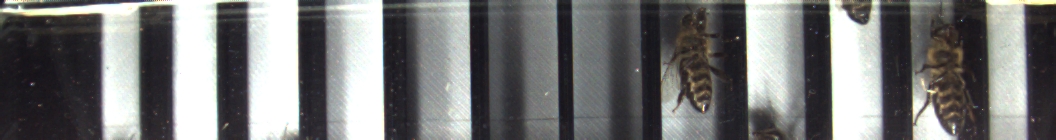}
        \caption{A typical sample from the BUT1 dataset. Note the disturbing background and improper illumination.}
    	\label{fig:BUT1}
    \end{subfigure}
    
    \begin{subfigure}[ht!]{0.9\textwidth}
        \centering
        \includegraphics[width=\textwidth]{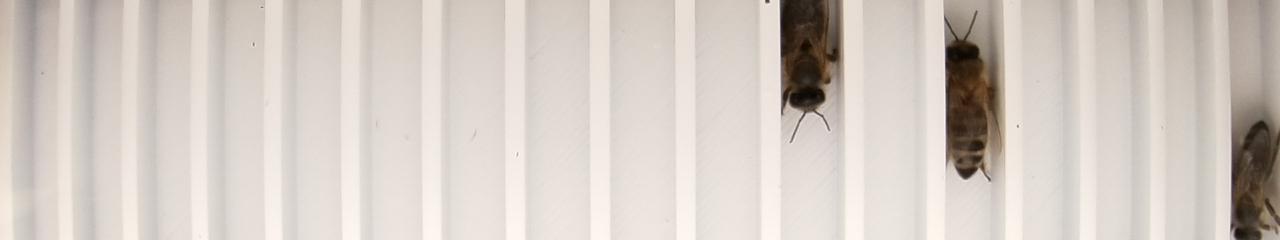}
        \caption{A typical sample from the BUT2 dataset. Note the homogeneous background and illumination.}
    	\label{fig:BUT2}
    \end{subfigure}
    
    \caption{An illustration of the samples from the BUT1 and BUT2 datasets.}
	\label{fig:BeeDS}
\end{figure*}

To collect these datasets, we used our device described in~\cite{Bilik2021}. Both datasets distinguish five classes - a complete bee heading in or out of the beehive, the head part, the abdominal part, and a cluster of several bees. Frames were captured with the frequency of 5 FPS, which ensures that every passing bee will be captured at least three times (head part - complete bee - abdominal part), and the annotations are in Yolo format. An illustration of both datasets is shown in Fig.~\ref{fig:BeeDS} with the measurement setup shown in Fig.~\ref{fig:MSetup}.

\subsection{Conventional computer vision-based approach}

As none of the previously mentioned conventional motion tracking techniques proved to be suitable for our needs, we deviate from analyzing the movement of individual objects of interest utilizing our prior knowledge of the environment and analyze only an overall movement in the region of interest.

To keep a reliable representation of our environment even after a longer period, we utilize a dynamic background model. We initialize our dynamic model $m(x,y,k)$ optionally with the first frame of the sequence or with a previously captured frame of the environment $f(x,y,k=0)$.

\begin{equation}
    m(x,y,0) = f(x,y,0)
\end{equation}

In each time step of the sequence we analyze the overall dynamic properties of the frame with thresholded subtraction

\begin{equation}
    d_1(x,y,k) = 
    \begin{cases}
        1  & \text{if } \left\lvert f(x,y,k) - f(x,y,k-1) \right\rvert > T_1 \\
        0  & \text{otherwise}
    \end{cases}
\end{equation}

where $T_1$ is the empirically selected threshold. And based on the overall number of dynamic pixels, we flag the scene as either dynamic or non-dynamic

\begin{equation}
    D_1(k) = \sum _{x,y}d_1(x, y, k) > T_2
\end{equation}

where $T_2$ is again the empirically selected threshold. Our dynamic model is updated only when the scene is flagged as static in sequence, meaning $D_1$ is flagged as \textit{false}, and the scene is flagged as static concerning the dynamic model. This flag is calculated similarly as the $D_1$ flag.

\begin{equation}
    d_2(x,y,k) = 
    \begin{cases}
        1  & \text{if } \left\lvert f(x,y,k) - m(x,y,k-1) \right\rvert > T_1 \\
        0  & \text{otherwise}
    \end{cases}
\end{equation}

\begin{equation}
    D_2(k) = \sum _{x,y}d_2(x, y, k) > T_2
\end{equation}

The dynamic model is then updated with a simple adaptive filter

\begin{equation}
    m(x, y, k) = 
    \begin{cases}
        \alpha \cdot f(x,y,k) + (1-\alpha)\cdot m(x, y, k-1)  \\ \text{if } D1(k) \text{ and } D2(k) \\ \\
        m(x, y, k-1)  \\ \text{otherwise}
    \end{cases}
\end{equation}

where $\alpha$ is the learning rate of the dynamic model.

To track the level of dynamic activity in our region of interest, we inspect the ratio of dynamically flagged pixels in the subtracted frame from the current time step with our dynamic environment model to the number of pixels in the region. However, this metric alone does not carry any information in terms of the movement's direction. To compensate for this, we split the inspected region into $N$ sections along the $y$ axis.

\begin{equation}
    \begin{gathered}
        f(x,y) \rightarrow g(x,y,n) \\
        \rm I\!R^{X\times Y} \rightarrow \rm I\!R^{X\times \left\lfloor \frac{Y}{N}\right\rfloor  \times N}
    \end{gathered}
\end{equation}

In each of these $N$ sections, we calculate the already mentioned metric of dynamic activity

\begin{equation}
    d(x,y,n,k) = 
    \begin{cases}
        1  & \text{if } \left\lvert f(x,y,n,k) - m(x,y,n,k) \right\rvert > T_1 \\
        0  & \text{otherwise}
    \end{cases}
\end{equation}

\begin{equation}
    r(n, k) = \frac{1}{X \cdot Y} \sum _{x,y} d(x,y,n,k)
\end{equation}

this $r(n,k)$ signal now carries enough information to determine the level and direction of movement in the observed region. To further ease this signal's processing, we approximate the first-order partial derivative concerning the time step dimension with differentiation

\begin{equation}
    dr(n, k) = r(n, k) - r(n, k-1)
\end{equation}

In the resulting signal $dr(n, k)$, we threshold its peaks to classify the current timestep with one of three classes: bee arrival ($class = 1$), idle state ($class = 0$), and bee departure ($class = -1$).

\begin{equation}
    class(n, k) = 
    \begin{cases}
        1  & \text{if } dr(n, k) > T_3 \\
        -1  & \text{if } dr(n, k) < -T_3 \\
        0  & \text{otherwise}
    \end{cases}
\end{equation}

where $T_3 \in \left\langle 0, 1\right\rangle $ is empirically selected threshold value.

To implement the actual counter of bees that traverse the region in one way or the other, we keep track of classes from the last $K_{max}$ time steps. We add a track to a list on each bee arrival, and with each bee departure, we flag an unflagged track as valid. We increment a counter once a valid track is present in all $N$ sections. The direction of the bee's movement is based on the age of the valid tracks on the region's edges.

As this is quite a simple approach, it's bound to have limitations; its main disadvantage is that if the bees do not travel independently but in packs, $r(n,k)$ will remain close to constant, $dr(n, k)$ will be close to zero. The bees won't be accounted for as $dr(n, k)$ does not cross the $T_3$ threshold value. Another limitation we have observed in experiments is that once a bee slows down at some point of its traverse through the observed area or the lightning in the observed area lowers its intensity, the $dr(n, k)$ values reduce, sometimes even to a noise level. This leads to a missing track entry in one or more sections of a tunnel, the bee is not accounted for and there may be hanging tracks left in sections where the bee was registered. This effect can have a detrimental effect on the next registered bees as hanging tracks on the edges of the observed region define the estimated traverse direction. This effect can be suppressed by lowering the maximum track age $K_{max}$, but lowering this value also effectively reduces sensitivity when the bee's velocity lowers. On the bright side, this approach runs under 20ms on Raspberry Pi 4B, sequentially processing 12 bee tunnels in each frame.

\subsection{CNN based approach}

For our experiments, we had decided to use the pre-trained SqueezeNet v1.1, ResNet-50 and DarkNet-53 network models. The built-in application of Matlab - Deep Network Designer was used. All these networks have been trained on more than a million images from the ImageNet database. They can classify images into 1000 categories of objects such as a keyboard, a coffee mug, a pencil, and many animals. As a result, the networks have learned rich feature representations for a wide range of images. ResNet-50 is a 50-layer CNN (48 convolutional layers, one MaxPool layer, and one average pool layer) with 25.6 million parameters and a model size of 96 MB. Darknet-53 has a similar number of layers (53 layers) but is roughly 1.6 times larger (41.6 M parameters, 155 MB model size).
In contrast, SqueezeNet is considerably smaller (18 layers deep, 1.2 M parameters, and 5.2 MB model size), yet achieves relatively high accuracy. In general, larger CNNs require more memory and computing power and consume more energy. It takes longer to train or retrain the network and classify a new image itself. Therefore, smaller networks are more feasible to deploy on edge devices with constrained resources, such as various embedded devices.

For all networks, the last convolutional and classification layers were adjusted to sort the images into five categories. Several combinations of network settings were always tested, mainly different mini-batch sizes, learning rates, solver, and number of epochs. The resulting retrained networks were applied to the test part of the datasets.

\subsection{Object detection based approach}

To evaluate the object detector approach we decided to test the current superstar in the neural network-based object detector field, the Yolov8 (You Only Look Once) architecture which is a evolution of the original Yolo \cite{yolo}. The Yolov8 provides the wide range of various architecture complexities. We decided to utilize the two extremes, the smallest and the largest ones, to test the miniature network that is easy to deploy on the edge device and the maximal approach that evaluates the maximal capabilities of the given architecture.

For each combination of architecture and dataset we trained network for 100 epochs, with dynamic batch size and tested results on a separated test data subset as shown in Table~\ref{tb:YoloSetup}. The training time was from few minutes in case of Yolov8n on BUT1 data, and about an hour and a half for Yolov8x on BUT2 data.

For this experiment, we utilize the official PyTorch-based Yolov8 Ultralytics library \cite{yolov8_ultralytics}, which provides a flexible and easy-to-use environment for model training and evaluation. The computer hardware configuration for this project included an NVIDIA RTX 3090 GPU with 24 GB memory complemented by an AMD Ryzen 9 processor and 64 GB of RAM, ensuring efficient data handling and preprocessing capabilities.

\begin{table}[ht!]
    \begin{center}
    \caption{Yolo training setup}\label{tb:YoloSetup}
    \begin{tabular}{ccccc}
    Dataset               & Yolov8 Arch & Batch size & Epoch Dur[s] & mAP[50] \\ 
    \hline
    \multirow{2}{*}{BUT1} & n           & 122        & 5         & 0.87   \\
                          & x           & 18         & 30        & 0.90   \\
    \hline
    \multirow{2}{*}{BUT2} & n           & 122        & 8         & 0.93   \\
                          & x           & 18         & 50        & 0.94   \\ 
    \hline
    \end{tabular}
    \end{center}
\end{table}

\section{Experimental results}

Experimental results for all investigated methods are shown in Table.~\ref{tb:ResultsAll}. The detection accuracies were computed as relative error of the total number of predictions in respect to the ground truth annotations.

It could be seen that the worst performed the conventional computer vision approach, which highly underestimates the number of bees in both directions and on both datasets. This is most likely caused by the bees coming through in clusters, which was not expected during the algorithm design, but proved to be very common during the performed measurements. The best detection accuracies were achieved using the ResNet-50 CNN classifier consistently on the both datasets. The object detection using both variants of the YoloV8 model did not fulfilled the expectations while achieving significantly worse results than the ResNet-50.

\begin{table*}[ht!]
    \begin{center}
    \caption{Experimental results. Best results for both datasets are shown in bold.}\label{tb:ResultsAll}
    \begin{tabular}{ccrrrrr}
                          &         & \multicolumn{1}{c}{Ground truth} & \multicolumn{1}{c}{CCV approach} & \multicolumn{1}{c}{ResNet-50} & \multicolumn{1}{c}{Yolov8 n} & \multicolumn{1}{c}{Yolov8 x} \\
    \midrule
                          
    \multirow{2}{*}{BUT1} & Bee in                & 182  & 29     & 179    & 127    & 127   \\
                          & Bee out               & 24   & 6      & 26     & 28     & 37    \\
                          
                          & Accuracy \textit{in}  & -    & 0.16   & \textbf{0.98}   & 0.69   & 0.69  \\
                          & Accuracy \textit{out} & -    & 0.25   & \textbf{0.92}   & 0.84   & 0.46  \\
    \midrule
    
    \multirow{2}{*}{BUT2} & Bee in                & 590  & 117    & 583    & 410    & 416   \\
                          & Bee out               & 1311 & 124    & 1249   & 967    & 1200  \\
                          
                          & Accuracy \textit{in}  & -    & 0.19   & \textbf{0.98}   & 0.69   & 0.70  \\
                          & Accuracy \textit{out} & -    & 0.09   & \textbf{0.95}   & 0.73   & 0.91  \\
    
    \hline
    
    \end{tabular}
    \end{center}
\end{table*}

\begin{figure}[ht!]
    \begin{center}
    \includegraphics[width=8.4cm]{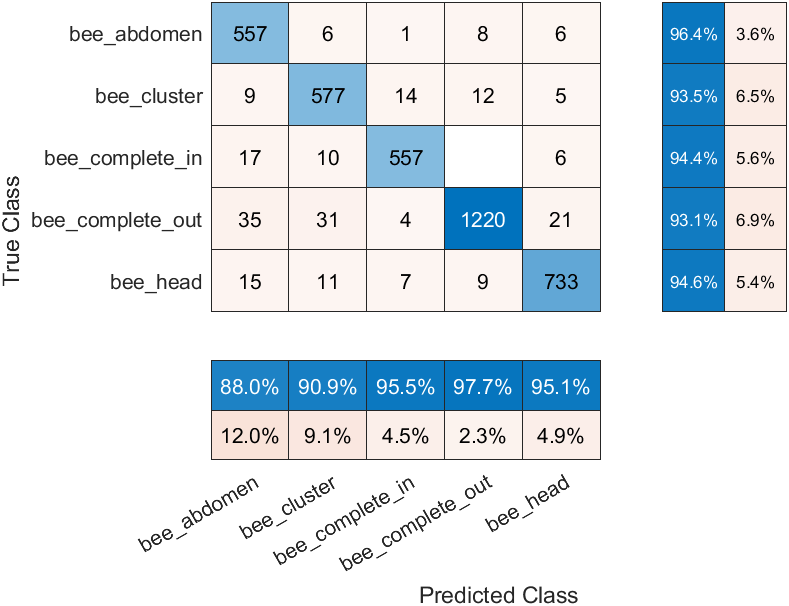}    
    \caption{Confusion matrix of the selected ResNet model trained over all classes.} 
    \label{fig:ConfMat_ResNet}
    \end{center}
\end{figure}

CNN classifier Resnet-50 was trained with the following parameters: batch size 15, learning rate 0.01, and SGDM solver. Fig.~\ref{fig:ConfMat_ResNet} shows the confusion matrix of this model determined on the test data of the BUT2 dataset. A higher classification accuracy was achieved on the BUT2 dataset (98\% versus 87.30\% on the BUT1 dataset), probably due to the overall higher number of images and the more even number of images in each category. If both datasets were experimentally combined, an accuracy of 93.22\% was achieved on all 4438 test image cutouts.

As expected, SqueezeNet performed worse (90.26\% on BUT2 and 84.48\% on BUT1), but given the significantly smaller size, the roughly 3-4\% drop in accuracy is not that significant. On the other hand, the larger DarkNet-53 network did not provide better results. Although it learned more stably with an accuracy greater than 93\%, it did not outperform the most successful ResNet-50 network model in any combination of parameters.

The YoloV8 object detector's variants did not perform consistently over the datasets - in the case of the BUT1 dataset, the YoloV8n performed better in the case of the outcoming accuracy (0.84 \% over 0.46 \%) and in the case of the BUT2 dataset, better results in both directions were achieved with the YoloV8x. From the embedded point of view, it might be more practical to use the smaller model YoloV8n as it has lower computational demands.

\section{Conclusions}

In this work, we tackle the problem of counting bees leaving and arriving at a hive. After a summary of motion-tracking and bee-counting techniques in the second chapter, we propose conventional computer vision, CNN classifier-based and object detector-based approaches to solve this problem in the third chapter.

The first proposed method is based on conventional computer vision techniques and did not perform as expected. Probably due to high sensitivity to the bees entering the hive close to each other in the form of a cluster or a row, the proposed method fails to distinguish individual bees, and therefore, it highly underestimates the number of bees entering and leaving the hive. Nevertheless, this method has a low computational demand and was successfully tested on the RaspberryPi 4 embedded computer. As the most important are the trends over the time, underestimating the number of bees might not be such critical if it is consistent. Therefore, we will focus on this analysis in our future research.

The CNN classifier-based approach performed the best of all three investigated methods using the ResNet-50 classifier. This approach achieved the highest accuracy and successfully counted almost the same number of bees as in the case of ground truth labels. It was also able to train on limited datasets, especially in the case of the BUT1. As our future work, we will focus on exploring a computational less demanding networks to be used for an automated bee counting module within our device under development.

The object detector approach did not performed as well as expected on both investigated datasets, when it performed significantly worse in comparison with the ResNet-50. This is probably caused by mismatching the bees if they are advancing not individually, but in the form of cluster, or a row as in the case of the conventional computer vision approach. The object detectors are also probably too complex and computationally demanding for this task.

The bee counting task is relatively complex due to a high similarity of bees in both orientations and due to the nature of their movement in our measurement system, where they usually touch each other, sometimes stand on place, or return after passing underneath the camera. Further accuracy improvement will also require a state machine, which will consider the other detections, such as the bee head or abdomen, to track the individual bees more precisely. Nevertheless, as said above, capturing a trend in the count of incoming and outcoming bees is sufficient for most of the apiary analysis.

\begin{ack}
The completion of this paper was made possible by grant No. FEKT-S-23-8451 - "Research on advanced methods and technologies in cybernetics, robotics, artificial intelligence, automation and measurement" financially supported by the Internal science fund of Brno University of Technology.
\end{ack}



\end{document}